\documentclass[journal]{IEEEtran}

\usepackage{graphicx}
\graphicspath{{figures/}}
\usepackage{float}
\usepackage{commath,amsmath,amssymb} 
\usepackage{tabularx}
\usepackage{color}
\usepackage{url}
\usepackage{array}
\usepackage{ragged2e}
\usepackage{todonotes}
\usepackage[numbers,sort&compress]{natbib}
\usepackage[scaled]{beramono}
\usepackage[T1]{fontenc}
\usepackage{makecell}

\newcommand{\actlbl}[1]{\texttt{\small #1}}
\def\vec#1{\mathchoice{\mbox{\boldmath$\displaystyle#1$}}
  {\mbox{\boldmath$\textstyle#1$}}
  {\mbox{\boldmath$\scriptstyle#1$}}
  {\mbox{\boldmath$\scriptscriptstyle#1$}}}

\begin{document}
%
\title{A Multi-viewpoint Outdoor Dataset for Human Action Recognition}
%
%
%


\author{Asanka G. Perera, Yee Wei Law, Titilayo T. Ogunwa, and Javaan Chahl,~\IEEEmembership{Member,~IEEE}%
}

%
%

%

\maketitle

\begin{abstract}
Advancements in deep neural networks have contributed to near perfect results for many computer vision problems such as object recognition, face recognition and pose estimation. However, human action recognition is still far from human-level performance. Owing to the articulated nature of the human body, it is challenging to detect an action from multiple viewpoints, particularly from an aerial viewpoint. This is further compounded by a scarcity of datasets that cover multiple viewpoints of actions. To fill this gap and enable research in wider application areas, we present a multi-viewpoint outdoor action recognition dataset collected from YouTube and our own drone. The dataset consists of 20 dynamic human action classes, 2324 video clips and 503086 frames. All videos are cropped and resized to 720$\times$720 without distorting the original aspect ratio of the human subjects in videos. This dataset should be useful to many research areas including action recognition, surveillance and situational awareness. We evaluated the dataset with a two-stream CNN architecture coupled with a recently proposed temporal pooling scheme called kernelized rank pooling that produces nonlinear feature subspace representations. The overall baseline action recognition accuracy is 74.0\%. 
\end{abstract}


%
\IEEEpeerreviewmaketitle

\section{Introduction}
%
%
%
%

\IEEEPARstart{I}{nterpretting} the dynamics of the human body in videos and inferring the action the human subject is performing is a challenging research problem. Sometimes, even for a human observer, it is difficult to accurately recognize actions in many scenarios. Such scenarios include scenes with occlusions, motion blur, low resolution, perspective distortion, etc. However, popular video datasets such as \cite{soomro12ucf101,kuhne11hmdb,jhuang13towards,zhang13actemes,karpathy14large,heilbron15activitynet,feichtenhofer16convolutional,kay17kinetics} have provided a stimulus to research in human action recognition, with many successes reported over the last few years based on deep learning techniques.

Owing to the articulated nature and infinite poses the human body can make, human action recognition is a challenging research problem, and is far from resolved. Recent advances in autonomous driving have created a lot of interest among researchers attempting to understand human actions using different data capturing methods. This includes analyzing challenging videos recorded from moving cameras \cite{geiger12kitti,keller11newbenchmark,zhang17citypersons}. There are a number of datasets available to study road environments especially focusing on the behavior of pedestrians and vehicles \cite{geiger12kitti,keller11newbenchmark,dollar12pedestrian,cordts16cityscapes,yu18BDD100K}. Research related to autonomous vehicles rely heavily on such datasets. 

In addition to autonomous ground vehicles such as self-driving cars and buses, there is increasing commercial and research interest in application areas such as search and rescue \cite{help17erdelj,human13peschel}, surveillance, crowd monitoring \cite{finn12unmanned,kaff17vbii}, sports activity recording \cite{kaff17vbii} and situational awareness. In oder to achieve human-level performance in such challenging missions a significant amount of training data and affective algorithms are needed.

To meet these demands, datasets that cover common outdoor actions from multiple viewpoints are necessary. Currently most available video datasets are limited to ground-based action recognition \emph{or} aerial action recognition. It is difficult to find a single dataset that contains clips from both aerial and ground viewpoints covering realistic outdoor actions. Our work presented here is specifically focused on providing a \emph{multi-viewpoint} and single/multi-person video dataset to the research community to advance research in outdoor human action recognition.

The dataset presented in this article consists of 20 action classes collected from YouTube and our own drone. Most of the actions with an aerial viewpoint were recorded from an aerial platform (UAV or helicopter) or a rooftop camera while the actions with ground-level viewpoints were recorded from a fixed or moving camera. This dataset is suitable for use on research involving action recognition, situational awareness and surveillance. All the videos were cropped around the action and resized to a resolution of 720$\times$720 without distorting the original aspect ratio of the action.

As the baseline action recognition algorithm, we used a two-stream CNN architecture coupled with a recently proposed temporal pooling scheme called \emph{kernelized rank pooling based on feature subspaces} (KRP-FS) \cite{cherian18nonlinear} that produces nonlinear spatiotemporal feature subspace representations. KRP-FS transforms nonlinear action dynamics to a low-rank subspace in the (potentially) infinite-dimensional reproducing kernelized Hilbert space (RKHS) \cite{scholkopf01learning}, and creates linear and efficient feature descriptor for action classification. The baseline accuracy of KPR-FS for our dataset was found to be comparable to those for recently published human action datasets.

The rest of this article is organized as follows. Section~\ref{sec_related_work} discusses closely related work on popular action recognition datasets and their limitations in terms of suitability for multi-viewpoint human action recognition. Section~\ref{sec_preparing_the_dataset} describes the steps involved in preparing the dataset, and compares the dataset with other recently published video datasets. Section~\ref{sec_experimental_results} reports experimental results. A discussion of issues and potential improvements is presented in Section~\ref{sec_discussion}. Section~\ref{sec_conclusion} concludes.

\section{Related work}\label{sec_related_work}

The surveys \cite{chaquet13survey} and \cite{kang16review} compare action recognition datasets that were published up to the years 2013 and 2016 respectively. Here, we discuss datasets related to our work.

\textbf{Early datasets}: The problem of recognizing human pose in statically captured videos has been studied extensively in recent literature~\cite{kang16review}. Early datasets such as KTH~\cite{schuldt04recognizing} and Weizmann~\cite{blank05actions} have helped progress action recognition research. However, the videos were captured indoor using a static camera. The static backgrounds used restrict their applicability in a variety of studies. One of the largest and most diverse datasets in this category is UCF101~\cite{soomro12ucf101}, with 13320 selected instances from YouTube videos. The videos belong to 101 action classes with dynamic camera movements and different backgrounds. However, the actions are mainly related to indoor and sports events with a low resolution of QVGA (320$\times$240). 

\textbf{Recent small datasets}: Recent datasets created from YouTube and movie clips include JHMDB \cite{jhuang13towards} and Penn action \cite{zhang13actemes}. Videos in these datasets have been annotated for both pose and action. JHMDB has 21 action classes across 928 videos (320x240 resolution) and Penn action consists of 15 actions across 2326 video sequences with a maximum dimension of 480 pixels (width or height). These datasets contain both indoor and outdoor scenes and static and dynamic camera movement. However, the types of action and limited outdoor scenes from a moving camera make them inappropriate for multi-viewpoint action recognition.

\textbf{Recent large datasets}: The datasets discussed so far are limited in size. Large action datasets such as Sports-1M~\cite{karpathy14large} and ActivityNet~\cite{heilbron15activitynet} were introduced in the recent past. They mainly focus on actions recorded from ground cameras and do not provide spatial localization details of the human subjects. For example, Sports-1M videos are annotated automatically using YouTube topics and are therefore, weakly labeled~\cite{feichtenhofer16convolutional}.

There are large action recognition datasets that were prepared from realistic YouTube videos. ActivityNet~\cite{heilbron15activitynet}, YouTube-8M \cite{abuelhaija16youtube8m} Kinetics \cite{kay17kinetics} and HACS \cite{zhao19hacs} are such popular datasets. The first release of Kinetics dataset \cite{kay17kinetics} consists of 400 action classes with 400 minutes of video. It covers a diverse range of actions collected from YouTube videos. The successive Kinetics releases have 600 and 700 classes respectively. The HACS \cite{zhao19hacs} dataset consists of 1.55~M annotated clips across 200 classes. The diversity of these large datasets helps to pre-train action recognition network on them and enhance the network performance on small datasets.

\textbf{Aerial action recognition datasets}: Of increasing interest to the action recognition community is aerial videos because drones are an increasingly popular means of capturing videos. Human action detection from an aerial view proves more challenging than from a fronto-parallel view. Oh et al.~\cite{oh11large} presented a large-scale VIRAT dataset with about 550 videos. The 29 hours of video in the dataset cover 23 types of events, recorded from both static and moving cameras. Using resolutions of 1920$\times$1080 and 1280$\times$720, the VIRAT ground dataset was recorded using stationary overhead cameras placed in multiple locations. However, in the case of the VIRAT aerial dataset, using a low resolution of 480$\times$720 increased the difficulty in retrieving rich activity information from human subjects that cover a relatively small part of the frame. Both the ground and aerial datasets were recorded against cluttered and uncontrolled backgrounds. 

The Okutama-Action 4k resolution video dataset introduced in \cite{barekatain17okutama} focused on concurrent detection of actions by multiple subjects. A clutter-free baseball field was where the videos were recorded using 2 UAVs. The abrupt camera movements captured 12 actions, taken from various view angles and at altitudes from 10 to 45 meters. The 90 degree elevation angle of the camera resulted in self-occlusions and severe perspective distortion in the videos. 

Other important aerial action datasets are Mini-drone~\cite{bonetto15privacy}, UCFARG~\cite{online11ucfarg} and UCF aerial action~\cite{online11ucfaerial}. An R/C-controlled blimp and a helium balloon were used to record videos for UCF aerial action and UCF ARG respectively with similar action classes. While the UCF ARG dataset is multi-viewpoint (recorded from aerial, rooftop and ground cameras), the UCF aerial action is a single-viewpoint dataset. The Mini-drone dataset was developed as a surveillance dataset in order to evaluate various aspects and definitions of privacy. The recording of the dataset was done in a car park with a low-altitude flying drone to detect actions categorized under illicit, suspicious and normal behaviors.

Research on outdoor action recognition would benefit from more datasets, such as the dataset presented here which covers 20 common outdoor actions captured from multiple viewpoints.

\section{Preparing the dataset}\label{sec_preparing_the_dataset}

This section explains the data collection process, the selected action classes and variations in the collected data. The dataset is compared with recently published aerial and ground video datasets.
  
\begin{figure*}
\centering
\includegraphics[width=0.8\textwidth]{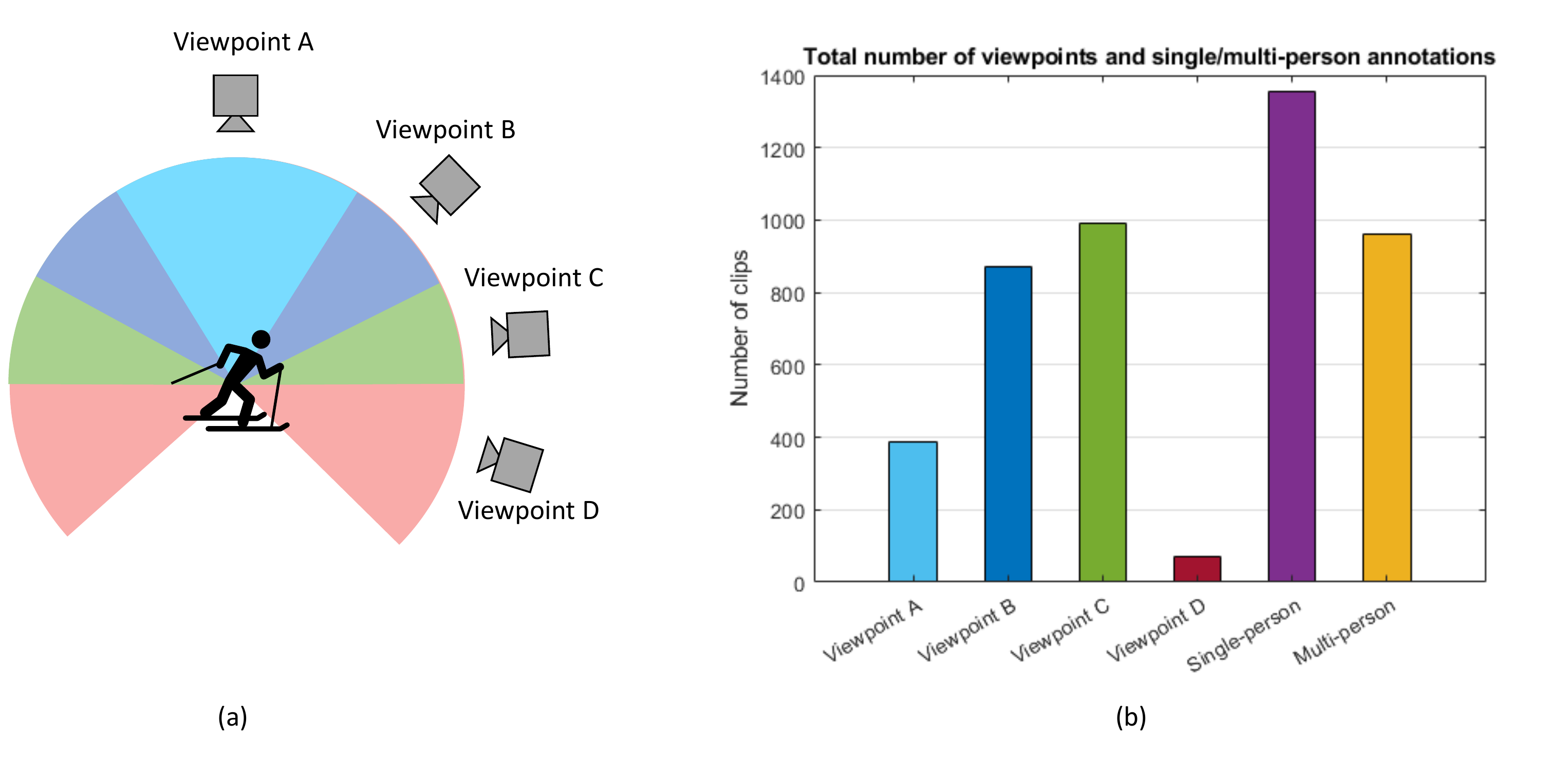}
\caption{Each clip was annotated with its viewpoint and number of people (single-person or multi-person). (a) The upper viewing hemisphere is discretized into three sections (viewpoints A, B and C) and any viewpoint falls into the lower viewing hemisphere is considered as viewpoint D. (b) Number of clips per viewpoint type and people count.}
\label{fig:annot_stats}
\end{figure*}

\subsection{Data collection}

We collected a total of 2324 video clips: 6 of them captured using our own quadrotor UAV, and 2318 from YouTube published under the Creative Commons license. 

All clips were resized to 720$\times$720 pixels. We cropped and resized the clips from their original videos without distorting their original aspect ratios. All of our clips have an undistorted $1:1$ aspect ratios making them easy to use as inputs to CNN architectures. All videos were re-sampled at 29.97 fps.

When collecting the data we tried to cover multiple viewpoints of the same actions while prioritizing aerial viewpoints. We discretized the viewing sphere of the action as shown in Fig.~\ref{fig:annot_stats}(a). The viewpoint distribution across the dataset is shown in Fig.~\ref{fig:annot_stats}(b). Among the viewpoints, $54\%$ are aerial viewpoints (viewpoints A and B) while $46\%$ are ground viewpoints (viewpoints C and D). Videos with aerial viewpoints were mostly recorded by an aerial platform (a UAV or helicopter) or rooftop cameras. The dataset contains videos recorded from both fixed and moving cameras. 

\begin{figure*}[ht!]
\centering
\includegraphics[width=\textwidth]{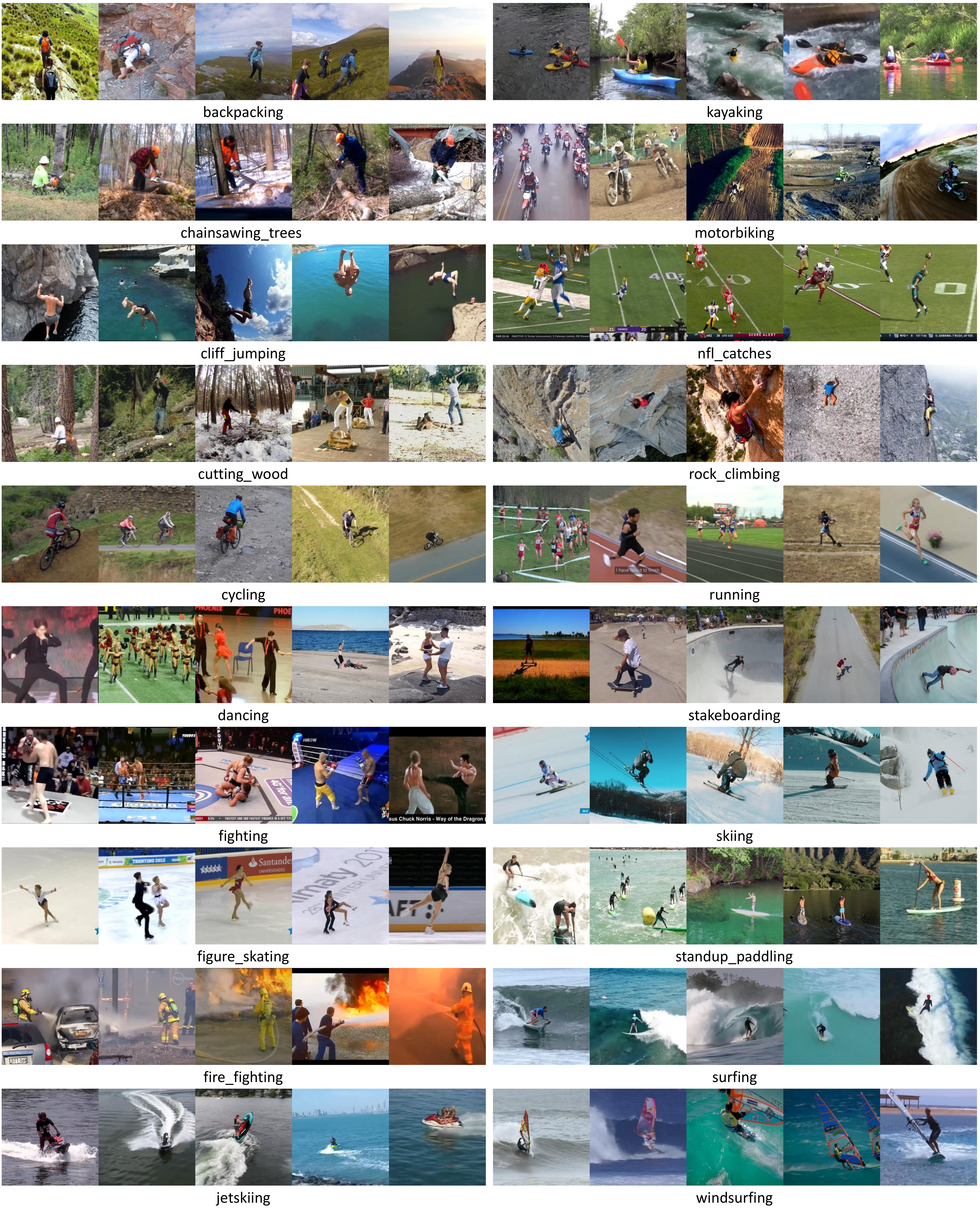}
\caption{Randomly selected images from the 20 classes.}
\label{fig:sample_images}
\end{figure*}

The collected videos cover realistic scenarios across 20 selected classes (see Figure \ref{fig:sample_images}). The 20 selected actions were \actlbl{backpacking}, \actlbl{chainsawing\_trees}, \actlbl{cliff\_jumping}, \actlbl{cutting\_wood}, \actlbl{cycling}, \actlbl{dancing}, \actlbl{fighting}, \actlbl{figure\_skating}, \actlbl{fire\_fighting}, \actlbl{jetskiing}, \actlbl{kayaking}, \actlbl{motorbiking}, \actlbl{nfl\_catches}, \actlbl{rock\_climbing}, \actlbl{running}, \actlbl{stakeboarding}, \actlbl{skiing}, \actlbl{standup\_paddling}, \actlbl{surfing} and \actlbl{windsurfing}. Since our focus was to create an outdoor action recognition dataset we selected 18 outdoor actions consisting of popular outdoor sports (e.g., \actlbl{jetskiing}, \actlbl{surfing}) and general outdoor events (e.g., \actlbl{chainsawing\_trees}, \actlbl{fire\_fighting}). The remaining two actions, namely \actlbl{dancing} and \actlbl{fighting}, are applicable to both indoor and outdoor settings.

There are substantial variations in the actions in terms of the phase, orientation, camera movement, viewpoint, number of humans in an action, human sizes and the backgrounds. These variations create a challenging dataset for action recognition, and also makes it more representative of real-world situations.

\subsection{Data annotation}
All clips were annotated for the action class, viewpoint and the number of people involved in the action. The number of people was annotated as ``single-person'' or ``multi-person''. 

\subsection{Dataset summary}
The dataset contains a total of 2324 clips with 503086 frames, consistently cropped to a resolution of 720$\times$720 and re-sampled at 29.97 fps. The dataset is 280 
minutes (i.e., 4 hours and 40 minutes) long, with an average clip length of 7.4 seconds. For every action class, there are a minimum of 105 clips, a maximum of 153 clips, and an average of 116.2 clips. Table~\ref{table:summary} provides a summary of the dataset. 

\begin{figure*}
  \centering \includegraphics[height=0.9\textheight]{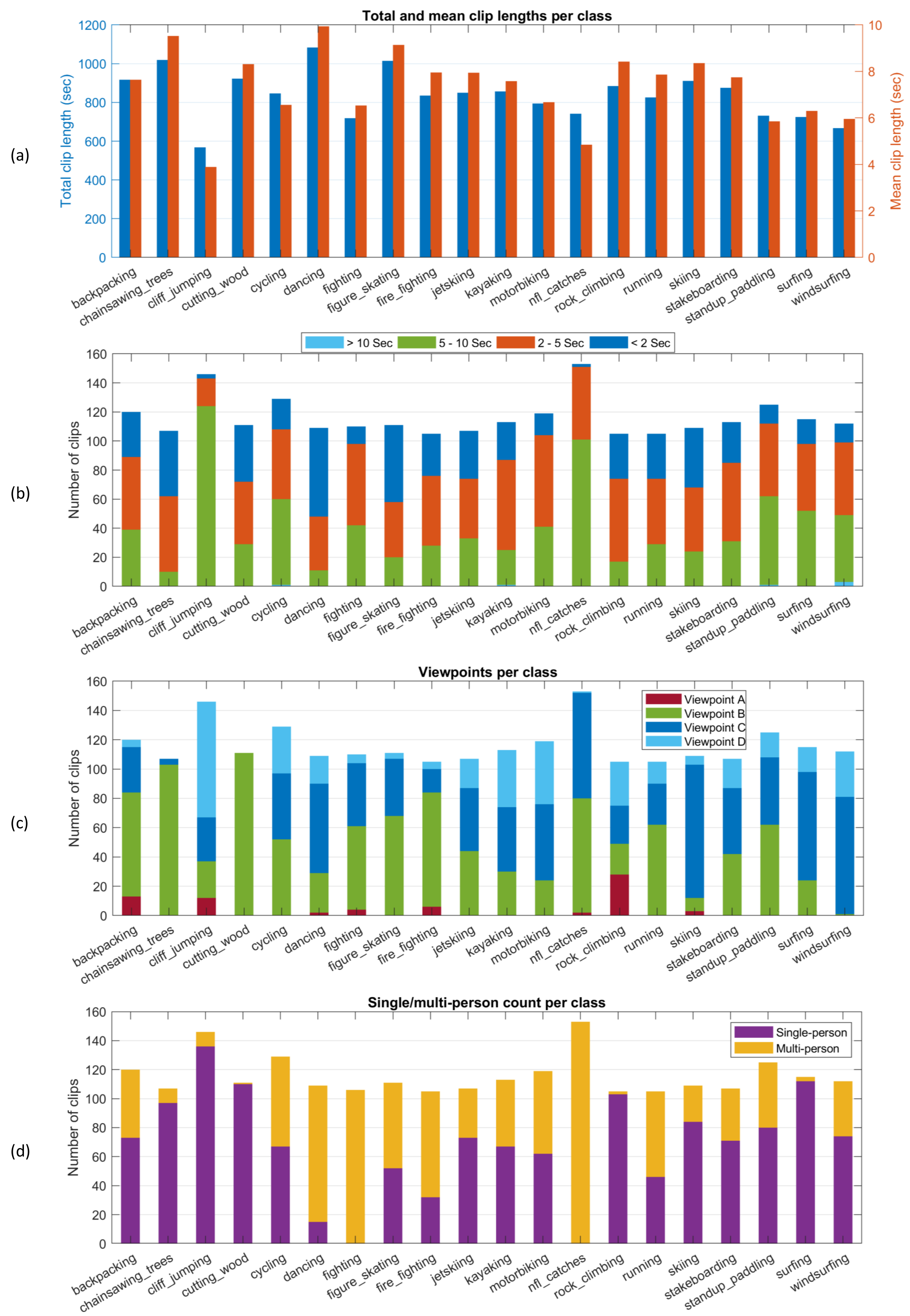}
  \caption{(a) Total and mean clip lengths per action class. (b) Clip lengths grouped per action class. (c) Viewpoint distribution per action class. (d) Distribution of single/multi-person clips per action class.}
  \label{fig:perclass_stats}
\end{figure*}

Figure~\ref{fig:perclass_stats}(a) shows the total clip length (blue) and mean clip length (amber) for each action class. Actions \actlbl{cliff\_jumping} and \actlbl{nfl\_catches} have a relatively low clip durations since both are very short actions. The clip durations are grouped and shown in Figure~\ref{fig:perclass_stats}(b). The longest clip duration is 21.3 seconds and the shortest is 1.84 seconds. The majority of the clips lie in the 2-10 seconds range.

Fig.~\ref{fig:perclass_stats}(c) shows the viewpoint breakdown across the classes. There is a low number of clips with viewpoint D. This is due to the nature of actions selected here. The \actlbl{rock\_climbing} action has the largest number of viewpoint D clips, because most of the \actlbl{rock\_climbing} actions were recorded by a person on the ground. Actions \actlbl{chansawing\_trees} and \actlbl{cutting\_wood} were mostly recorded from the ground-level viewpoint C. In contrast, the \actlbl{windsurfing} action has the largest number of aerial viewpoints owing to the aerial camera platforms used to record them. The other 16 actions have a relatively balanced aerial and ground viewpoint distribution.

Figure~\ref{fig:perclass_stats}(d) shows the distribution of single-person versus multi-person clips per action class. There is a mix of single-person and multi-person clips for all classes excepting \actlbl{cutting\_wood}, \actlbl{fighting} and \actlbl{nfl\_catches}. \actlbl{cutting\_wood} is normally a single-person action, and \actlbl{fighting} and \actlbl{nfl\_catches} are always multi-person actions.

\begin{table}
  \centering
  \caption{A summary of the dataset.}\label{table:summary}
  \begin{tabular}{|l|l|}
  \hline
  Feature & Value\\
  \hline
  \# Actions            &   20\\
  \# People             &   One to many\\
  \# Clips              &   2324\\
  \# Clips per class    &   116.2\\
  Mean clip length      &   7.4 sec\\
  Total duration        &   280 mins\\
  \# Frames             &   503086\\
  Frame rate            &   29.97 fps\\
  Resolution            &   720$\times$720\\
  Viewpoints         	  &   Multi-viewpoints\\
  Annotations           &   Action, single/multi-person, viewpoint\\
  Sources 			        &   YouTube and own drone\\
  \hline
  \end{tabular}
\end{table}

\subsection{Comparison with other datasets}

In Table~\ref{table:compare}, we compare our dataset with some recently published video datasets. These datasets have helped to advance research in action recognition and event recognition. Datasets such as UCF101, HMDB51 and JHMDB have been used as the standards in  recent action recognition studies. The design of our dataset is similar to these datasets, but there are some unique differences in our dataset. We selected a range of mostly outdoor actions and collected videos to cover multiple viewpoints. Further, the resolution of our dataset is 720$\times$720 compared to the 360$\times$240 resolution of UCF101, HMDB51 and JHMDB.

Drone-recorded datasets like VIRAT, Mini-drone and Okutama-Action are focused on various degrees of outdoor action recognition. Compared to these datasets, our dataset covers a more diverse range of outdoor actions. Recently released Kinetics\footnote{\url{https://deepmind.com/research/open-source/open-source-datasets/kinetics/}} and HACS\footnote{\url{http://hacs.csail.mit.edu/}} datasets support large numbers of classes, and provide download scripts and annotations for a large number of YouTube videos, but since these videos are online, they may disappear over time. In comparison, ours is an offline dataset\footnote{\url{https://asankagp.github.io/mod20/}} and are nonvolatile.

\begin{table*}
  \centering
  \caption{Comparison of recently published video datasets.}\label{table:compare}
  \begin{tabular}{|l|l|l|l|l|l|l|l|}
  \hline
  Dataset & Scenario & Purpose & Environment & Frames & Classes & Resolution & Year\\
  \hline
  UT Interaction \cite{ryoo09spation}       &   Surveillance        &   Action recognition    &   Outdoor &   36k   &   6   &   360$\times$240    &   2010\\
  VIRAT \cite{oh11large}                    &   Drone, surveillance &   Event recognition     &   Outdoor &   Many  &   23  &   Varying           &   2011\\
  HMDB51 \cite{kuhne11hmdb}             	  &   Movies, YouTube     &   Action recognition    &   Varying &   655k  &   51  &   320$\times$240    &   2011\\
  UCF101 \cite{soomro12ucf101}              &   YouTube             &   Action recognition    &   Varying &   558k  &   24  &   320$\times$240    &   2012\\
  J-HMDB \cite{jhuang13towards}             &   Movies, YouTube     &   Action recognition    &   Varying &   32k   &   21  &   320$\times$240    &   2013\\
  Mini-drone \cite{bonetto15privacy}        &   Drone               &   Privacy protection    &   Outdoor &   23.3k &   3   &   1920$\times$1080  &   2015\\
  Okutama-Action~\cite{barekatain17okutama} &   Drone               &   Action recognition    &   Outdoor &   70k   &   13  &   3840$\times$2160  &   2017\\
  Kinetics \cite{kay17kinetics}				      &   YouTube				      & 	Action recognition 	  &   Varying &   Many  &   400 &   Varying           &   2017\\
  HACS \cite{zhao19hacs}					          &   YouTube				      &   Action recognition    &   Varying &   Many  &   200 &   Varying           &   2019\\
  MOD20 (this dataset)                      &   YouTube, own drone  &   Action recognition    &   Varying &   503k  &   20  &   720$\times$720    &   2019\\
  \hline
  \end{tabular}
\end{table*}

\subsection{Generation of training and test sets}
We experimented with the dataset by generating 3 split sets, with a $70:30$ train-to-test ratio for each split set. After separating the clips into classes, the first, mid and last 30\% of clips for each action class were assigned to test sets 1, 2 and 3 respectively, while the remainder were assigned to the corresponding training sets. This split generation method avoided or minimized the use of clips generated from the same original video in both train and test splits.


\section{Experimental Results}\label{sec_experimental_results}

We evaluated our dataset using the metric of classification accuracy, calculated using the scores returned by the action classifiers. 

For action classification, we used a two-stream CNN architecture coupled with a recently proposed temporal pooling scheme called \emph{kernelized rank pooling based on feature subspaces} (KRP-FS) \cite{cherian18nonlinear} that produces nonlinear spatiotemporal feature subspace representations, as our baseline algorithm. KRP-FS was selected because it has been shown to produce state-of-the-art results for several action recognition datasets such as JHMDB, HMDB51, MPII Cooking Activities and UT-Kinect actions. We used the publicly available KRP-FS code\footnote{\url{https://github.com/anoopcherian/kernelized_rank_pool/}} and fine-tuned it for our dataset. 

Our evaluation results include results for KRP-FS as well as for \emph{basic kernelized rank pooling} (BKRP), as a means of sanity check. Valid results should show that KRP-FS provides better accuracy than BKRP.

\subsection{Introduction to BKRP and KRP-FS}

The feature representation employed here is based on \emph{rank pooling}. Rank pooling refers to the temporal pooling strategy where a learning-to-rank algorithm \cite{liu09learning} is trained on different videos of the same action, and the parameters of the resultant ``ranking machine'' are used as a video representation that captures the video-wide temporal evolution of the video \cite{fernando15modeling}.

The first ever rank pooling scheme to be proposed is \emph{VideoDarwin} \cite{fernando15modeling}. Denote an $n$-frame video by
\begin{equation}X_n = [\vec{x}_1, \vec{x}_2, \ldots, \vec{x}_n],\end{equation}
where $\vec{x}_t\in\mathbb{R}^d$ is the $d$-dimensional feature vector of the frame at time $t$. VideoDarwin
\begin{itemize}
  \item defines a vector-valued function $V:\{X_t,t\}\to\vec{v}_t\in\mathbb{R}^d$;
  \item imposes the order constrains $\vec{v}_n\succ\cdots\succ\vec{v}_t\succ\cdots\succ\vec{v}_1$;
  \item trains a ranking machine to learn a \emph{linear} function characterized by parameters $\vec{u}\in\mathbb{R}^D$, namely $\phi(\vec{v};\vec{u})=\vec{u}^\top\vec{v}$, such that $\vec{v}_t\succ\vec{v}_s\iff\vec{u}^\top\vec{v}_t>\vec{u}^\top\vec{v}_s$.
\end{itemize}
The parameter vector $\vec{u}$ is essentially a compact representation of $X$ that summarizes the sequence dynamics. Defining $\vec{v}_t\triangleq\vec{m}_t/\|\vec{m}_t\|$, where $\vec{m}_t\triangleq(\vec{x}_1+\cdots+\vec{x}_t)/t$, gives us the variant of VideoDarwin called the \emph{Forward VideoDarwin}.


Observing that VideoDarwin uses linear projection --- a poor fit for nonlinear geometries --- Cherian et al. \cite{cherian18nonlinear} proposed applying the kernel trick \cite{scholkopf01learning} to rank pooling, enabling $X$ to be mapped to a potentially infinite-dimensional \emph{reproducing kernel Hilbert space} (RKHS), in which data is linear. Suppose for kernel $K(\vec{x}_t,\vec{z})$, there exists a feature map $\Phi:\mathbb{R}^d\times\mathbb{R}^d\to\mathcal{H}$, where $\mathcal{H}$ is a Hilbert space for which $\langle\Phi(\vec{x}_t),\Phi(\vec{z})\rangle_\mathcal{H}=K(\vec{x}_t,\vec{z})$. The most basic form of kernelized rank pooling is the \emph{basic kernelized rank pooling} (BKRP) formulation, which finds the pre-image $\vec{z}$ whose embedding $\Phi(\vec{z})$ in the feature space defines a direction/line, onto which the embeddings $\Phi(\vec{x}_1),\ldots,\Phi(\vec{x}_n)$ are projected in a way that preserves the temporal order. The BKRP scheme uses the pre-image $\vec{z}$ as the action description.



BKRP assumes that the pre-image $\vec{z}$ always exists. However, given that pre-images are finite-dimensional representatives of infinite-dimensional Hilbert space points, they may not be unique or may not even exist \cite{mika99kernel}. As a solution, instead of estimating a single hyperplane in feature space, the \emph{kernelized rank pooling based on feature subspaces} (KRP-FS) formulation estimates a low-rank kernelized subspace subject to the constraint that projections of the embeddings $\Phi(\vec{x}_1),\ldots,\Phi(\vec{x}_n)$ onto this subspace should preserve the temporal order (as increasing distances from the subspace origin). Both BKRP and KRP-FS assume nonlinear input data and allow compact linear order-preserving projections.

\subsection{Feature extraction using KRP-FS}

A two-stream CNN architecture --- the same as that of P-CNN \cite{cheron15pcnn} --- was used, where one stream extracts RGB features from RGB images, and the other optical flow features from optical flow images. For extracting RGB features, the ImageNet-trained ``VGG-f'' network provided by Chatfield et al. \cite{chatfield14return} was used. For extracting (optical) flow features, the motion network that Gkioxari and Malik \cite{gkioxari15finding} trained on the UCF101 split 1 dataset was used. Both networks have the same architecture consisting of 5 convolutional layers and 3 fully-connected layers, as shown in Table~\ref{table:cnn}. The CNN features were obtained from the $7^{th}$ fully connected layer of both networks. The loss functions of RGB and flow networks are \emph{dropout} and \emph{cross entropy} respectively. RGB and flow data were used for kernelized rank pooling computations.

\begin{table*}
  \centering
  \caption{The two-stream CNN architecture employed: Each cell represents a convolutional layer $(C(n \times k \times k))$ or fully connected layer $(FC(n))$ with $n$ filters and kernel size $k \times k$.}\label{table:cnn}
  \begin{tabularx}{\linewidth}{|p{2cm}|X|X|X|X|X|p{1.2cm}|p{1.2cm}|p{1.2cm}|}
  \hline
  Network 							& C1	& C2  & C3 & C4	& C5 &  FC6 & FC7 & FC8 \\
  \hline
  RGB \cite{chatfield14return} (input size: $224 \times 224$)      &  $C(64\times11\times11)$ & $C(256\times5\times5)$ & $C(256\times3\times3)$ & $C(256\times3\times3)$ & $C(256\times3\times3)$ & $FC(4096)$ (dropout) & $FC(4096)$ (dropout) & $FC(1000)$ (softmax)\\
  \hline
  Flow \cite{gkioxari15finding} (input size: $227 \times 227$)    &  $C(96\times7\times7)$ & $C(384\times5\times5)$ & $C(512\times3\times3)$ & $C(512\times3\times3)$ & $C(384\times3\times3)$ & $FC(4096)$ (cross entropy) & $FC(4096)$ (cross entropy) & $FC(1000)$ (softmax)\\
  \hline
  \end{tabularx}
\end{table*}

\begin{figure*}
  \centering
  \includegraphics[width=\textwidth]{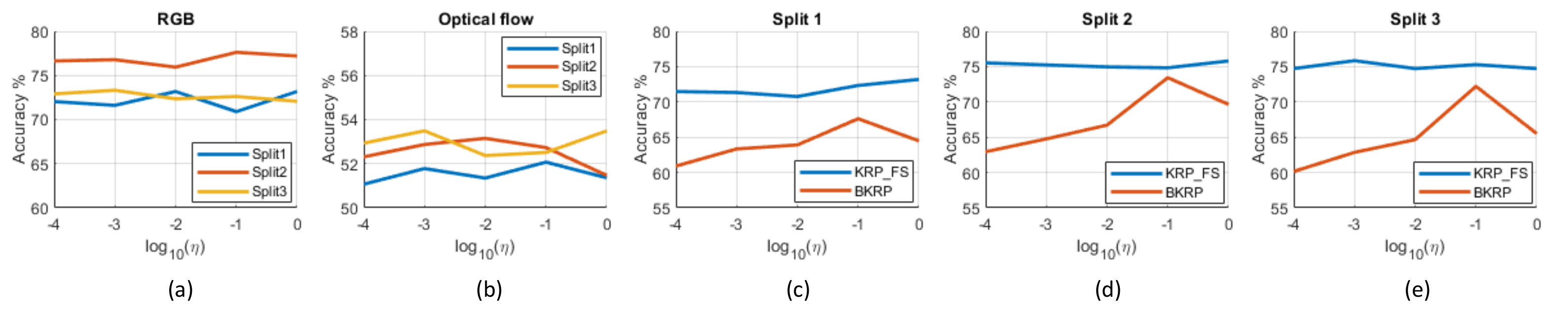}
  \caption{(a) Average accuracies of KRP-FS using RGB features for different ranking threshold values. (b) Average accuracies of KRP-FS using flow features for different rank threshold values. (c)-(e) Average accuracies of KRP-FS and BKRP for the split sets 1, 2 and 3.}
  \label{fig:line_graphs}
\end{figure*}

Here, we briefly discuss two important parameters for feature extraction using KRP-FS: the ranking threshold $\eta$ and the Nystr\"{o}m (approximation) ratio.

The ranking threshold $\eta$ is the parameter that enforces the temporal order of rank pooling calculation. $\eta$ can be used to verify if an algorithm reliably detects temporal order in the extracted features. In Fig.~\ref{fig:line_graphs}, we analyzed the influence of $\eta$ on the classification accuracy by varying it from 0.0001 to 1  at a multiplicative step size of 10. Fig.~\ref{fig:line_graphs}(a) and (b) show the accuracy variation with different thresholds for RGB and flow data respectively. Fig.~\ref{fig:line_graphs}(c)-(e) show the average accuracies of KRP-FS and BKRP for the 3 splits of our dataset. KRP-FS achieves the highest average accuracy when $\eta=1$. 

Nystr\"{o}m kernel approximation \cite{drineas05nystrom} is a technique for reducing the computational cost of kernel matrix calculation. It computes only a few columns of the kernel matrix and approximates the full kernel by a low-rank outer product. Table~\ref{table:nystrom} shows the classification accuracies for five Nystr\"{o}m ratios calculated on split 1 of our dataset. The data was uniformly sampled in the order of $(1/2^j)^{th}$ the original data size, while $j$ is varied from 5 to 2 at a step size of $-1$. Consistent with \cite{cherian18nonlinear}, $(1/8)^{th}$ was found to be the optimal Nystr\"{o}m ratio for the HMDB51 dataset. We experimented with different Nystr\"{o}m values and found that $(1/8)^{th}$ was the optimal Nystr\"{o}m ratio for our dataset too. 

\begin{table}
  \centering
  \caption{Analyzing Nystr\"{o}m approximation ratios.}\label{table:nystrom}
  \begin{tabular}{|c|c|}
  \hline
  Nystr\"{o}m approx. ratio 	& Accuracy (\%)\\
  \hline
	1/32            &   67.33\\
	1/16            &   68.61\\
	1/8             &   71.47\\
	1/4             &   71.04\\
  \hline
  \end{tabular}
\end{table}

\subsection{Performance evaluation}

Over 3 split sets, the average classification accuracy is 74.05\%. 

We used a 3.6 GHz four-core computer for the experiments. The average execution time (total time/frames) to achieve the baseline results was 1.43 milliseconds for KRP-FS (excluding the CNN forward pass time of 27.8 milliseconds per frame), consistent with Cherian et al.'s timing \cite{cherian18nonlinear}.

Table~\ref{table:performance} compares the accuracy of KRP-FS to that of BKRP for different action recognition datasets. The average classification accuracies of KRP-FS for our dataset, HMDB51, MPII Cooking and JHMDB lie in a close range (between 69\% and 74\%). The results can potentially be improved by using Fisher vector (IDT-FV) features \cite{peng14action} with KRP-FS \cite{cherian18nonlinear}.

\begin{table}
  \centering
  \caption{Comparison of the average accuracies (\%) of BKRP and KRP-FS for different datasets.}\label{table:performance}
  \begin{tabularx}{0.9\linewidth}{|p{1.7cm}|>{\centering\arraybackslash}X|>{\centering\arraybackslash}X|>{\centering\arraybackslash}X|>{\centering\arraybackslash}X|}

  \hline 
  Dataset 							& Rank Pooling \cite{fernando15modeling} 	& BKRP   	&  KRP-FS   & KRP-FS + IDT-FV \cite{cherian18nonlinear} \\
  \hline \hline
  UT-Kinect \newline actions \cite{xia12view}       &   75.5        			& 84.8  	&  99.0 	& -\\
  \hline
  HMDB51 \cite{kuhne11hmdb} \newline             	&   - 						& 64.1     	&  69.8 	& 72.7\\
  \hline
  MPII Cooking \cite{rohrbach12database} 			&   72.0     				& 66.3    	&  70.0 	& 76.1\\
  \hline
  JHMDB \cite{jhuang13towards} \newline            	&   -             			& 71.5    	&  73.8     & 74.2\\
  \hline
  MOD20 (this work)             					&   -    					& 66.55     &  74.0     & -\\
  \hline
  \end{tabularx}
\end{table}

\section{Discussion}\label{sec_discussion}

The actions collected for this work consists of challenging sequences from complex outdoor scenarios. There are videos recorded from moving aerial and ground cameras adding a rich motion variation to the dataset. We provide action class, viewpoint and person-count annotations with the dataset. We plan to release person bounding box annotations in the future.

We selected KRP-FS method to evaluate the dataset. This method has shown state-of-the-art results in several action recognition datasets. It does not use extra data for the computation and the run time can be optimized by selecting dataset specific parameters.

The classification accuracy of our dataset lies in the accuracy ranges of relatively similar action recognition datasets (UCF101, HMDB51, JHMDB  etc.). Since our dataset covers a unique multi-viewpoint outdoor action set, we believe that this work will be a valuable addition to the research community, particularly for studies focused on low-altitude aerial action recognition or outdoor multi-viewpoint action recognition.

This dataset has been created to help progress research in action recognition. The work presented in this paper is intended to provide a sufficient amount of data to cover some common outdoor human actions. We collected and annotated a large number of aerial videos (54\%) of actions to achieve a substantial multi-viewpoint video collection. These actions can be visible from a low-altitude slow-flying UAV and is hence suitable for low-altitude multi-viewpoint action recognition studies.

\section{Conclusion}\label{sec_conclusion}

We presented an action recognition dataset, called MOD20, consisting of videos collected from YouTube and our own drone. The dataset contains 2324 videos lasting a total of 240 minutes. The dataset was labeled with classes associated with 20 common outdoor human actions. The actions were selected from challenging and complex scenarios, and cover multiple viewpoints, from ground-level to bird's-eye view. The substantial variation in body size, number of people, viewpoints, camera motion, and background makes our dataset challenging for action recognition. The dataset contains videos recorded from both ground and aerial as well as stationary and moving cameras. The action classes, 720$\times$720 size un-distorted clips and multi-viewpoint video selection extend the dataset's applicability to a wider research community.

We evaluated this new dataset using KRP-FS feature descriptors, and reported an overall baseline action recognition accuracy of 74.0\%. This dataset is useful for research involving action recognition, activity recognition, and actor detection. The dataset is available at \url{https://asankagp.github.io/mod20/}.

\section*{Acknowledgment}

This project was assisted by Project Tyche, the Trusted Autonomy Initiative of the Defence Science and Technology Group (grant number myIP6780). We thank Anoop Cherian for his help with our kernelized rank pooling implementation.

\ifCLASSOPTIONcaptionsoff
  \newpage
\fi



%

\bibliographystyle{IEEEtran}
\bibliography{ref}

\begin{thebibliography}{10}
\providecommand{\url}[1]{#1}
\csname url@samestyle\endcsname
\providecommand{\newblock}{\relax}
\providecommand{\bibinfo}[2]{#2}
\providecommand{\BIBentrySTDinterwordspacing}{\spaceskip=0pt\relax}
\providecommand{\BIBentryALTinterwordstretchfactor}{4}
\providecommand{\BIBentryALTinterwordspacing}{\spaceskip=\fontdimen2\font plus
\BIBentryALTinterwordstretchfactor\fontdimen3\font minus
  \fontdimen4\font\relax}
\providecommand{\BIBforeignlanguage}[2]{{%
\expandafter\ifx\csname l@#1\endcsname\relax
\typeout{** WARNING: IEEEtran.bst: No hyphenation pattern has been}%
\typeout{** loaded for the language `#1'. Using the pattern for}%
\typeout{** the default language instead.}%
\else
\language=\csname l@#1\endcsname
\fi
#2}}
\providecommand{\BIBdecl}{\relax}
\BIBdecl

\bibitem{soomro12ucf101}
K.~Soomro, A.~R. Zamir, and M.~Shah, ``{UCF101}: {A} dataset of 101 human
  actions classes from videos in the wild,'' UCF Center for Research in
  Computer Vision, Tech. Rep., 2012.

\bibitem{kuhne11hmdb}
H.~Kuehne, H.~Jhuang, E.~Garrote, T.~Poggio, and T.~Serre, ``{HMDB}: A large
  video database for human motion recognition,'' in \emph{2011 International
  Conference on Computer Vision}, Nov 2011, pp. 2556--2563.

\bibitem{jhuang13towards}
H.~Jhuang, J.~Gall, S.~Zuffi, C.~Schmid, and M.~J. Black, ``Towards
  understanding action recognition,'' in \emph{2013 IEEE International
  Conference on Computer Vision}, Dec 2013, pp. 3192--3199.

\bibitem{zhang13actemes}
W.~Zhang, M.~Zhu, and K.~G. Derpanis, ``From actemes to action: A
  strongly-supervised representation for detailed action understanding,'' in
  \emph{2013 IEEE International Conference on Computer Vision}, Dec 2013, pp.
  2248--2255.

\bibitem{karpathy14large}
A.~Karpathy, G.~Toderici, S.~Shetty, T.~Leung, R.~Sukthankar, and L.~Fei-Fei,
  ``Large-scale video classification with convolutional neural networks,'' in
  \emph{2014 IEEE Conference on Computer Vision and Pattern Recognition}, June
  2014, pp. 1725--1732.

\bibitem{heilbron15activitynet}
F.~C. Heilbron, V.~Escorcia, B.~Ghanem, and J.~C. Niebles, ``{ActivityNet}: A
  large-scale video benchmark for human activity understanding,'' in \emph{2015
  IEEE Conference on Computer Vision and Pattern Recognition (CVPR)}, June
  2015, pp. 961--970.

\bibitem{feichtenhofer16convolutional}
C.~Feichtenhofer, A.~Pinz, and A.~Zisserman, ``Convolutional two-stream network
  fusion for video action recognition,'' in \emph{2016 IEEE Conference on
  Computer Vision and Pattern Recognition (CVPR)}, June 2016, pp. 1933--1941.

\bibitem{kay17kinetics}
W.~Kay, J.~Carreira, K.~Simonyan, B.~Zhang, C.~Hillier, S.~Vijayanarasimhan,
  F.~Viola, T.~Green, T.~Back, A.~Natsev, M.~Suleyman, and A.~Zisserman, ``The
  kinetics human action video dataset,'' \emph{CoRR}, vol. abs/1705.06950,
  2017.

\bibitem{geiger12kitti}
A.~Geiger, P.~Lenz, and R.~Urtasun, ``{Are we ready for autonomous driving? The
  KITTI vision benchmark suite},'' in \emph{2012 IEEE Conference on Computer
  Vision and Pattern Recognition}, June 2012, pp. 3354--3361.

\bibitem{keller11newbenchmark}
C.~G. Keller, M.~Enzweiler, and D.~M. Gavrila, ``A new benchmark for
  stereo-based pedestrian detection,'' in \emph{2011 IEEE Intelligent Vehicles
  Symposium (IV)}, June 2011, pp. 691--696.

\bibitem{zhang17citypersons}
S.~Zhang, R.~Benenson, and B.~Schiele, ``{CityPersons}: A diverse dataset for
  pedestrian detection,'' in \emph{2017 IEEE Conference on Computer Vision and
  Pattern Recognition (CVPR)}, July 2017, pp. 4457--4465.

\bibitem{dollar12pedestrian}
P.~Doll\'ar, C.~Wojek, B.~Schiele, and P.~Perona, ``Pedestrian detection: An
  evaluation of the state of the art,'' \emph{IEEE Transactions on Pattern
  Analysis and Machine Intelligence}, vol.~34, 2012.

\bibitem{cordts16cityscapes}
M.~Cordts, M.~Omran, S.~Ramos, T.~Rehfeld, M.~Enzweiler, R.~Benenson,
  U.~Franke, S.~Roth, and B.~Schiele, ``The cityscapes dataset for semantic
  urban scene understanding,'' in \emph{The IEEE Conference on Computer Vision
  and Pattern Recognition (CVPR)}, June 2016.

\bibitem{yu18BDD100K}
F.~Yu, W.~Xian, Y.~Chen, F.~Liu, M.~Liao, V.~Madhavan, and T.~Darrell,
  ``{BDD100K}: A diverse driving video database with scalable annotation
  tooling,'' \emph{CoRR}, vol. abs/1805.04687, 2018.

\bibitem{help17erdelj}
\BIBentryALTinterwordspacing
M.~Erdelj, E.~Natalizio, K.~R. Chowdhury, and I.~F. Akyildiz, ``Help from the
  sky: Leveraging uavs for disaster management,'' \emph{IEEE Pervasive
  Computing}, vol.~16, no.~1, pp. 24--32, Jan.-Mar. 2017. [Online]. Available:
  \url{doi.ieeecomputersociety.org/10.1109/MPRV.2017.11}
\BIBentrySTDinterwordspacing

\bibitem{human13peschel}
J.~M. Peschel and R.~R. Murphy, ``On the human–machine interaction of
  unmanned aerial system mission specialists,'' \emph{IEEE Transactions on
  Human-Machine Systems}, vol.~43, no.~1, pp. 53--62, Jan 2013.

\bibitem{finn12unmanned}
\BIBentryALTinterwordspacing
R.~L. Finn and D.~Wright, ``Unmanned aircraft systems: Surveillance, ethics and
  privacy in civil applications,'' \emph{Computer Law and Security Review},
  vol.~28, no.~2, pp. 184 -- 194, 2012. [Online]. Available:
  \url{http://www.sciencedirect.com/science/article/pii/S0267364912000234}
\BIBentrySTDinterwordspacing

\bibitem{kaff17vbii}
A.~Al-Kaff, F.~M. Moreno, L.~J.~S. Jos{\'e}, F.~Garc{\'i}a, D.~Mart{\'i}n,
  A.~de~la Escalera, A.~Nieva, and J.~L.~M. Garc{\'e}a, ``{VBII-UAV}:
  Vision-based infrastructure inspection-{UAV},'' in \emph{Recent Advances in
  Information Systems and Technologies}, {\'A}.~Rocha, A.~M. Correia, H.~Adeli,
  L.~P. Reis, and S.~Costanzo, Eds.\hskip 1em plus 0.5em minus 0.4em\relax
  Cham: Springer International Publishing, 2017, pp. 221--231.

\bibitem{cherian18nonlinear}
A.~Cherian, S.~Sra, S.~Gould, and R.~Hartley, ``Non-linear temporal subspace
  representations for activity recognition,'' in \emph{The IEEE Conference on
  Computer Vision and Pattern Recognition (CVPR)}, June 2018.

\bibitem{scholkopf01learning}
B.~Scholkopf and A.~J. Smola, \emph{Learning with Kernels: Support Vector
  Machines, Regularization, Optimization, and Beyond}.\hskip 1em plus 0.5em
  minus 0.4em\relax Cambridge, MA, USA: MIT Press, 2001.

\bibitem{chaquet13survey}
\BIBentryALTinterwordspacing
J.~M. Chaquet, E.~J. Carmona, and A.~Fernández-Caballero, ``A survey of video
  datasets for human action and activity recognition,'' \emph{Computer Vision
  and Image Understanding}, vol. 117, no.~6, pp. 633 -- 659, 2013. [Online].
  Available:
  \url{http://www.sciencedirect.com/science/article/pii/S1077314213000295}
\BIBentrySTDinterwordspacing

\bibitem{kang16review}
\BIBentryALTinterwordspacing
S.~Kang and R.~P. Wildes, ``Review of action recognition and detection
  methods,'' \emph{CoRR}, vol. abs/1610.06906, 2016. [Online]. Available:
  \url{http://arxiv.org/abs/1610.06906}
\BIBentrySTDinterwordspacing

\bibitem{schuldt04recognizing}
C.~Schuldt, I.~Laptev, and B.~Caputo, ``Recognizing human actions: a local svm
  approach,'' in \emph{Proceedings of the 17th International Conference on
  Pattern Recognition, 2004. ICPR 2004.}, vol.~3, Aug 2004, pp. 32--36 Vol.3.

\bibitem{blank05actions}
M.~Blank, L.~Gorelick, E.~Shechtman, M.~Irani, and R.~Basri, ``Actions as
  space-time shapes,'' in \emph{Tenth IEEE International Conference on Computer
  Vision (ICCV'05) Volume 1}, vol.~2, Oct 2005, pp. 1395--1402 Vol. 2.

\bibitem{abuelhaija16youtube8m}
S.~Abu-El-Haija, N.~Kothari, J.~Lee, A.~Natsev, G.~Toderici, B.~Varadarajan,
  and S.~Vijayanarasimhan, ``{YouTube-8M}: A large-scale video classification
  benchmark,'' \emph{CoRR}, vol. abs/1609.08675, 2016.

\bibitem{zhao19hacs}
H.~Zhao, Z.~Yan, L.~Torresani, and A.~Torralba, ``{HACS}: Human action clips
  and segments dataset for recognition and temporal localization,'' \emph{arXiv
  preprint arXiv:1712.09374}, 2019.

\bibitem{oh11large}
S.~Oh, A.~Hoogs, A.~Perera, N.~Cuntoor, C.~C. Chen, J.~T. Lee, S.~Mukherjee,
  J.~K. Aggarwal, H.~Lee, L.~Davis, E.~Swears, X.~Wang, Q.~Ji, K.~Reddy,
  M.~Shah, C.~Vondrick, H.~Pirsiavash, D.~Ramanan, J.~Yuen, A.~Torralba,
  B.~Song, A.~Fong, A.~Roy-Chowdhury, and M.~Desai, ``A large-scale benchmark
  dataset for event recognition in surveillance video,'' in \emph{CVPR 2011},
  June 2011, pp. 3153--3160.

\bibitem{barekatain17okutama}
M.~Barekatain, M.~Martí, H.~F. Shih, S.~Murray, K.~Nakayama, Y.~Matsuo, and
  H.~Prendinger, ``{Okutama-Action}: An aerial view video dataset for
  concurrent human action detection,'' in \emph{2017 IEEE Conference on
  Computer Vision and Pattern Recognition Workshops (CVPRW)}, July 2017, pp.
  2153--2160.

\bibitem{bonetto15privacy}
M.~Bonetto, P.~Korshunov, G.~Ramponi, and T.~Ebrahimi, ``Privacy in mini-drone
  based video surveillance,'' in \emph{2015 11th IEEE International Conference
  and Workshops on Automatic Face and Gesture Recognition (FG)}, vol.~04, May
  2015, pp. 1--6.

\bibitem{online11ucfarg}
{University of Central Florida}, ``{UCF-ARG} data set,''
  \url{http://crcv.ucf.edu/data/UCF-ARG.php}, November 2011.

\bibitem{online11ucfaerial}
------, ``{UCF aerial action dataset},''
  \url{http://crcv.ucf.edu/data/UCF_Aerial_Action.php}, November 2011.

\bibitem{ryoo09spation}
M.~S. Ryoo and J.~K. Aggarwal, ``Spatio-temporal relationship match: Video
  structure comparison for recognition of complex human activities,'' in
  \emph{2009 IEEE 12th International Conference on Computer Vision}, Sept 2009,
  pp. 1593--1600.

\bibitem{liu09learning}
T.-Y. Liu, ``Learning to rank for information retrieval,'' \emph{Foundations
  and Trends{\textregistered} in Information Retrieval}, vol.~3, no.~3, pp.
  225--331, 2009.

\bibitem{fernando15modeling}
B.~Fernando, E.~Gavves, J.~M. Oramas, A.~Ghodrati, and T.~Tuytelaars,
  ``Modeling video evolution for action recognition,'' in \emph{The IEEE
  Conference on Computer Vision and Pattern Recognition (CVPR)}, June 2015.

\bibitem{mika99kernel}
\BIBentryALTinterwordspacing
S.~Mika, B.~Sch\"{o}lkopf, A.~Smola, K.-R. M\"{u}ller, M.~Scholz, and
  G.~R\"{a}tsch, ``{Kernel PCA} and de-noising in feature spaces,'' in
  \emph{Proceedings of the 1998 Conference on Advances in Neural Information
  Processing Systems II}.\hskip 1em plus 0.5em minus 0.4em\relax Cambridge, MA,
  USA: MIT Press, 1999, pp. 536--542. [Online]. Available:
  \url{http://dl.acm.org/citation.cfm?id=340534.340729}
\BIBentrySTDinterwordspacing

\bibitem{cheron15pcnn}
G.~Cheron, I.~Laptev, and C.~Schmid, ``{P-CNN}: Pose-based {CNN} features for
  action recognition,'' in \emph{The IEEE International Conference on Computer
  Vision (ICCV)}, December 2015.

\bibitem{chatfield14return}
\BIBentryALTinterwordspacing
K.~Chatfield, K.~Simonyan, A.~Vedaldi, and A.~Zisserman, ``Return of the devil
  in the details: Delving deep into convolutional nets,'' \emph{CoRR}, vol.
  abs/1405.3531, 2014. [Online]. Available:
  \url{http://arxiv.org/abs/1405.3531}
\BIBentrySTDinterwordspacing

\bibitem{gkioxari15finding}
G.~Gkioxari and J.~Malik, ``Finding action tubes,'' in \emph{The IEEE
  Conference on Computer Vision and Pattern Recognition (CVPR)}, June 2015.

\bibitem{drineas05nystrom}
P.~Drineas and M.~W. Mahoney, ``On the nystr{\"o}m method for approximating a
  gram matrix for improved kernel-based learning,'' \emph{journal of machine
  learning research}, vol.~6, no. Dec, pp. 2153--2175, 2005.

\bibitem{peng14action}
X.~Peng, C.~Zou, Y.~Qiao, and Q.~Peng, ``Action recognition with stacked fisher
  vectors,'' in \emph{Computer Vision -- ECCV 2014}, D.~Fleet, T.~Pajdla,
  B.~Schiele, and T.~Tuytelaars, Eds.\hskip 1em plus 0.5em minus 0.4em\relax
  Cham: Springer International Publishing, 2014, pp. 581--595.

\bibitem{xia12view}
L.~{Xia}, C.~{Chen}, and J.~K. {Aggarwal}, ``View invariant human action
  recognition using histograms of 3d joints,'' in \emph{2012 IEEE Computer
  Society Conference on Computer Vision and Pattern Recognition Workshops},
  June 2012, pp. 20--27.

\bibitem{rohrbach12database}
M.~Rohrbach, S.~Amin, M.~Andriluka, and B.~Schiele, ``A database for fine
  grained activity detection of cooking activities,'' in \emph{2012 IEEE
  Conference on Computer Vision and Pattern Recognition}, June 2012, pp.
  1194--1201.

\end{thebibliography}

\end{document}